\documentclass{isprs}
\usepackage{subfigure}
\usepackage{setspace}
\usepackage[usenames, dvipsnames]{color}
\usepackage{tabu}
\usepackage[normalem]{ulem}
\usepackage[round]{natbib}

\newcommand{\myFigRef}[1]{fig.~(\ref{#1})}

\usepackage{geometry} 
\usepackage[utf8]{inputenc}
\usepackage{fancyhdr}
 
\begin{document}

\title{Deep Built-Structure Counting In Satellite Imagery Using Attention Based Re-weighting}

\author{
 Anza  Shakeel, 
 Waqas Sultani,
 Mohsen Ali\thanks{Corresponding author. This is useful to know for communication with the appropriate person in cases with more than one author}
 }

\address
{
    Information Technology University, Lahore, Pakistan \\(mscs15043, waqas.sultani, mohsen.ali)@itu.edu.pk\\
}


\abstract
{
In this paper, we attempt to address the challenging problem of counting built-structures in the satellite imagery.
Building density is a more accurate estimate of the population density, urban area expansion and its impact on the environment, than the built-up area segmentation. 
However, building shape variances, overlapping boundaries, and variant densities make this a complex task. 
To tackle this difficult problem, we propose a deep learning based regression technique for counting built-structures in satellite imagery. 
Our proposed framework intelligently combines features from different regions of satellite image using attention based re-weighting techniques.
Multiple parallel convolutional networks are designed to capture information at different granulates. 
These features are combined into the FusionNet which is trained to weigh features from different granularity differently, allowing us to predict a precise building count. 
To train and evaluate the proposed method, we put forward a new large-scale and challenging built-structure-count dataset. Our dataset is constructed by collecting satellite imagery from diverse geographical areas (planes, urban centers, deserts, etc.,) across the globe (Asia, Europe, North America, and Africa) and captures the wide density of built structures.
Detailed experimental results and analysis validate the proposed technique.
FusionNet has Mean Absolute Error of 3.65 and R-squared measure of 88\% over the testing data. Finally, we perform the test on the  $274.3 \times 10^{3}$ $m^2$ of the unseen region, with the error of 19 buildings off the 656 buildings in that area.  The dataset is available at http://im.itu.edu.pk/deepcount/.

}


\maketitle

\section{Introduction}\label{sec:Introduction}

Accurate, detailed and up-to-date analysis of the urban and non-urban areas play a vital role in building an economic and social understanding of the region, helping in policy making and designing interventions.
This analysis is dependent upon reliable and up-to-date surveys, which are lacking in the economically challenged areas of the world \citep{SatellitePovertyScienceJean16}.
 One of the important, but laborious to gather, statistics are population densities and built-up area, especially in either densely constructed areas or scarcely populated areas. An accurate and up-to-date mapping of the built-up areas is necessary for the effective disaster (e.g, flood or earthquake) relief, urban food security, and estimation of effects of the urbanization on the farmlands, forest volume, and population. Recently, where there has been a surge in using machine learning and satellite imagery to discover the economic and social pattern such as poverty \citep{SatellitePovertyScienceJean16}, slavery \citep{boyd2018slavery}, population spread, and large-scale urban patterns \citep{albert2017using}, there have been some successes in built-up-area estimation and building detection \citep{Facebookzhang2017building}. 
Unlike before mentioned works that rely on the collective features of the image to regress on the value, building detection requires detailed visual analysis, more accurately labeled data and respectable-resolution imagery. Over the years accurate results for the prediction of land-use and land-cover maps, such as 
 \citep{PatternNet2018,
 Langkvist2016, albert2017using} have been presented. 
However, either these are image classification based approaches or techniques that are restricted to just segmenting out the areas without coming up with a realistic count of structures. Furthermore, these approaches are not able to capture changes inside the urban regions. 

 
 Counting allows fine-grain urban population analysis and detailed view of change occurring within the urban and rural centers, without explicitly tackling the complex task of the individual building segmentation.
It is a better surrogate for the population analysis, more helpful in disaster management (damage and destruction estimation), urban food security analysis, and allow complex economic analysis of different parts of the city (indirectly allowing us to understand how much land is being used by each building). 

Several datasets have been introduced by different researches for various remote sensing applications. Available satellite imagery datasets include \citep{imgret6257473}, \citep{RSSCN7-7272047}, \citep{AID_Dataset-7907303}, \citep{ISPRSD}, \citep{xview} and \citep{PatternNet2018} comprising of land-use type class labels, bounding-boxes for overhead object localization and segmentation masks for categories like road, vegetation and buildings. Note that these land-use and land-cover datasets cover regions where buildings are separated from each other, or use hyper or multispectral imagery. Although these datasets are challenging and useful, none of them addresses the important problem of building counting in satellite imagery, especially in congested regions. 
\begin{figure*}[t!]
\begin{center}
        \includegraphics[width=15cm, height=9cm]{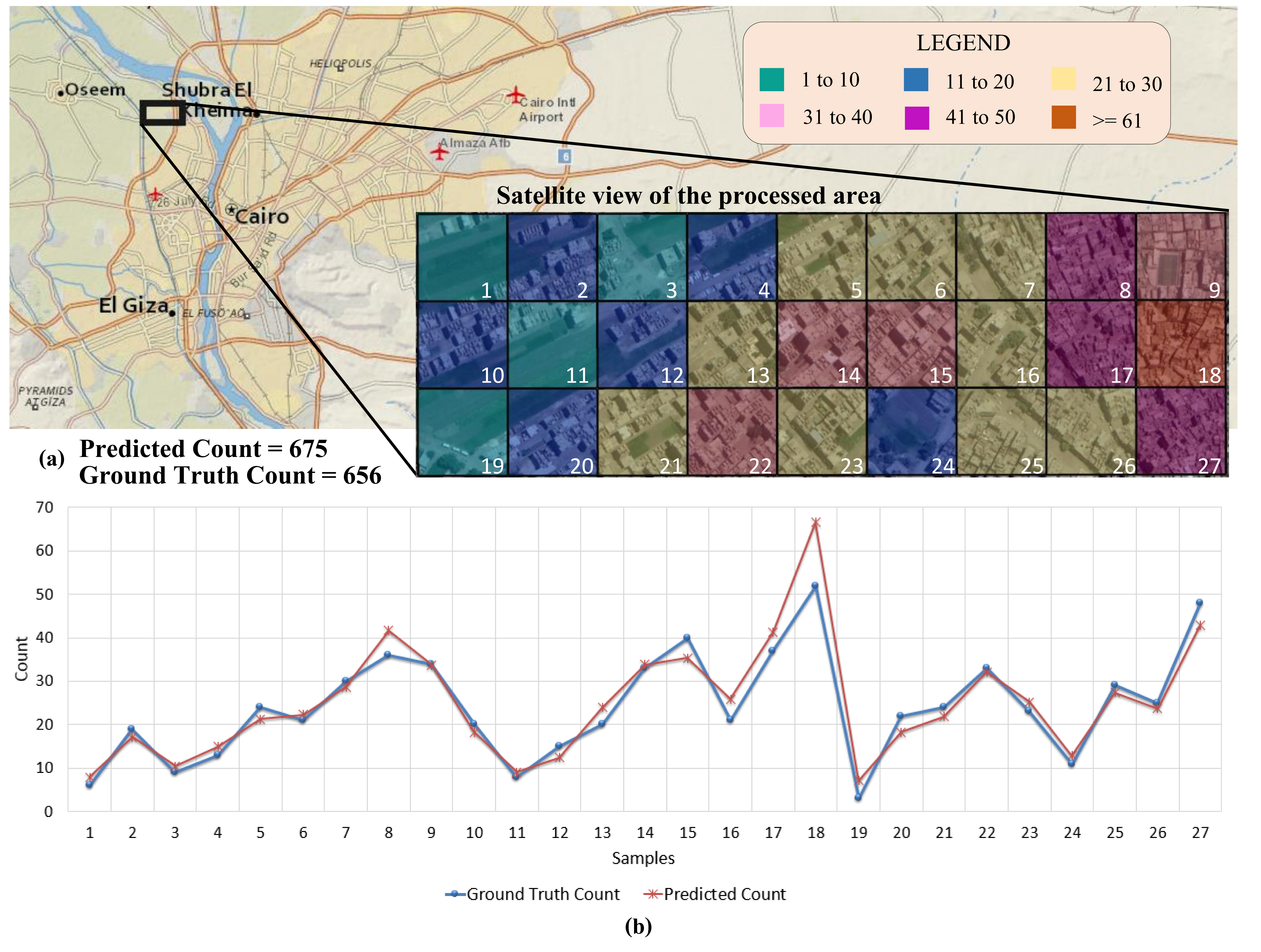}
    \caption{FusionNet: Built-structure count results. \textbf{(a)} $302.4 m \times 907.2 m$ area of Egypt is processed and the results are represented in the form of a heat map. The image is divided into 27 cells each of size $336\times336$. The predicted count is assigned a range of count. \textbf{(b)} Ground truth count in blue and predicted count in red for each cell is plotted. The precise building counts results demonstrate the generalization and robustness of our proposed deep model (Best viewed in color).}
    \label{large_image}
\end{center}
\end{figure*}
We venture into estimating the density of the buildings in the visible spectrum satellite imagery and present our counting results on the diverse set of images taken from sparsely to densely populated areas across the globe.
We propose a deep regression based network and two new attention based re-weighting techniques to achieve building counts. 
To do a thorough evaluation of our proposed approaches, we have collected a new large dataset of satellite imagery capturing built-structures of different densities (low, medium, and high) as well as including scenes without any built structure.  
Furthermore, we have provided detailed annotations for building counts for each satellite image.
To our knowledge, it is the first time challenging task of building counting has been handled at this scale. 
Our work exploits recent developments in the Deep Learning \citep{lecun2015deep} and propose the Convolutional Neural Network \citep{krizhevsky2012imagenet} based solution for estimating the number of buildings in the region.  
In summary, our work has the following technical contributions. 
\begin{enumerate}
    \item {We propose three new convolutional neural network based approaches for building counting. Firstly, we propose deep regression based counting. Secondly, we propose to employ attention network and introduce two new attention based re-weighting techniques to count the number of buildings.}
    \item {We propose large, diverse satellite imagery-based dataset with the hand-counted number of buildings.}
    \item {Extensive experiments to evaluate different approaches are performed. Experimental results demonstrate that our approach achieves state-of-art results as compared to competitive baselines.}
    \item Since we automatically estimate the built-regions through attention networks, We require only the image level count information, unlike previous methods, \citep{palmtreecount, switching-8099912, Learntorank},  which require sub-image level information .

\end{enumerate}

   \begin{figure}[h]
\begin{center}
        \includegraphics[width=1.0\columnwidth]{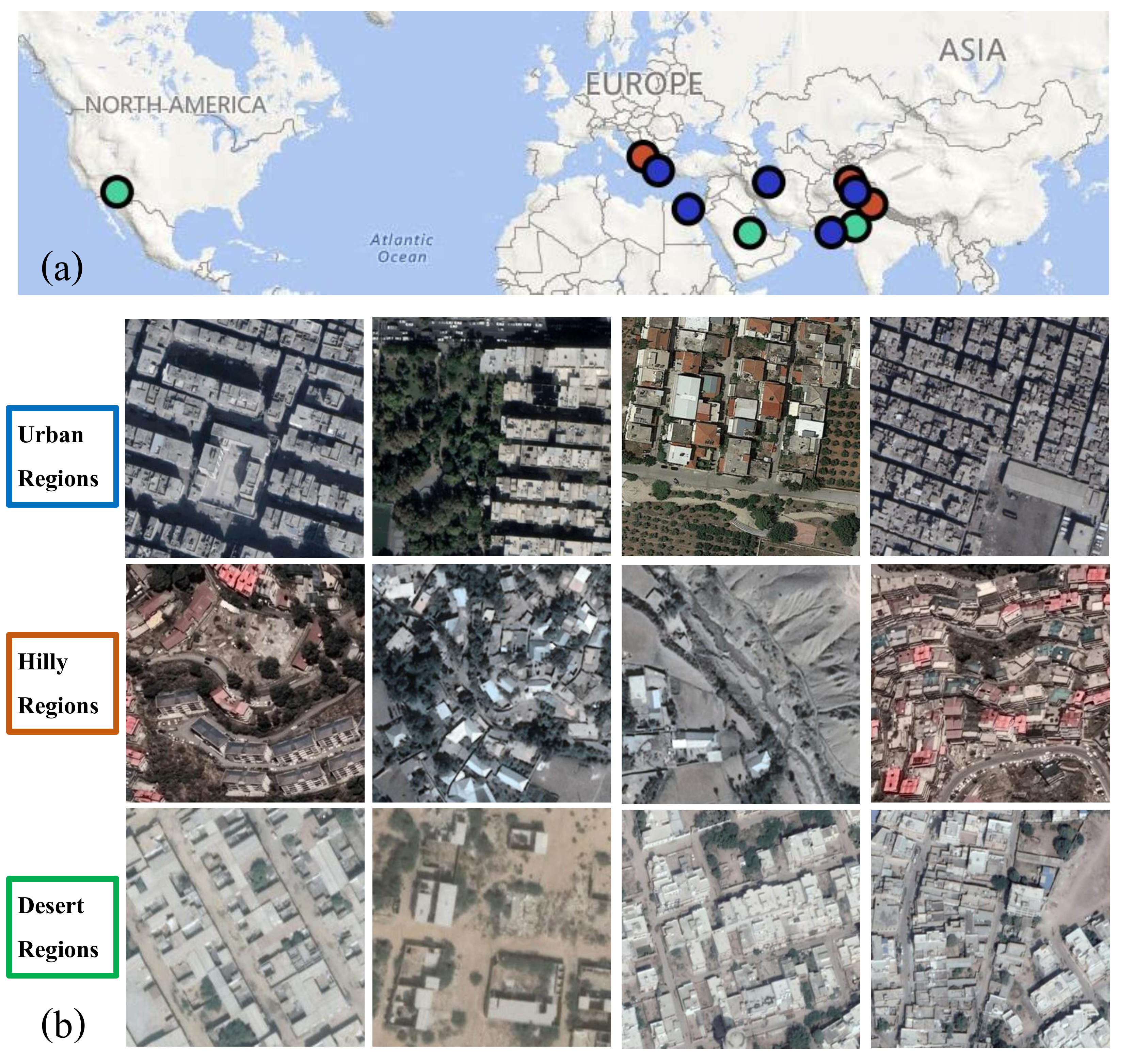}
    \caption{Dataset samples, (a) shows the geographic locations of the area of interests from where satellite imagery is taken. (b) presents the examples of urban, desert and hilly regions from our dataset. 
}\label{fig:Geographic_locations}
\end{center}
\end{figure} 

In what follows, we first provide a detailed review of existing techniques, details of our dataset collection is shared in Sec. \ref{sec:DataCollection} and in Sec. \ref{sec:Methodology} we present our proposed methodologies along with detailed implementation details.Sec. \ref{sec:ResultsandAnalysis} consists of results and their analysis. Finally Sec.~\ref{Conclusion} concludes the paper.

 
 
 
 





\section{Related Work}

Identifying urban markers in satellite imagery has been explored extensively. Most of these work differ in the quality of imagery, sensor, resolution of the imagery and the granularity at which results are reported.
The authors \cite{humanDevelopment} used high-resolution satellite imagery and night time satellite imagery as input to train CNN based deep neural network for predicting the economical markers representing human development. 
 \cite{damage-detection} proposed trees-of-shapes features to perform the damage detection using both the pre and post event satellite imagery.
 \cite{lalonde2018clusternet} performed object detection in the wide area motion imagery. { \cite{LearningRotationInvariant,LearningRotationInvariant_TIP} proposed the rotation invariant CNN for object detection in VHR remote sensing images. Note that since we automatically estimate the density built-regions through attention networks, we require only the image level count information instead of expensive object level annotations.}
 Similarly, there is extensive literature on building detection mainly relying on multi-spectral imaging. 
 Built-up area detection and building detection systems vary on the basis of the information they use (visible spectrum, multi-spectral imaging, DEM (Digital Elevation Model), LiDAR), features they extract (lines, corners, texture, etc,), machine learning applied on these features, the resolution of the input and output and the final objective they achieve. 
 Global Human Settlement Layer \citep{GHSLpesaresi2016operating} has been constructed using the Landsat imagery of multiple years, giving the percentage of built-up coverage in each pixel (38.2 m spatial resolution). 
Deep Learning based Semantic Segmentation \citep{FCN-7298965,SegNet-7803544} have been applied to the satellite imagery for land-use and land-cover analysis \citep{audebert2016semantic}.
Also, \cite{class_landcover_landuse} modified \citep{SegNet-7803544} to design a two-stage CNN that first segments land-cover type and then the segmented land-cover polygons are further processed for land-use classification. A boundary detector based semantic segmentation model is trained by \cite{edge_det_seg} and in this model, Digital Elevation Model (DEM) is used with the input image to train the pipeline. 
There is, however as per our knowledge, no previous work on counting the buildings from satellite imagery, let alone using the RGB spectrum.

Initial works relied on detecting the local features, such as edge, lines, corners
\citep{huertas1988detecting,  sirmacek2009urbanCornersSift}, or photo-metric properties \citep{muller2005robust,Ghaffarian2014}  and then intelligently combining these information together \citep{izadi2012three, krishnamachari1996delineating}. 
\cite{izadi2012three} looked for intersection of the lines and shadow cues to define the buildings. 
\cite{muller2005robust} used the low-level features to capture the geometric (roundness, size), photometric and structural properties (shadow, and presence of other houses that is neighborhood). 
They assumed that roof-tops are more visible in the red channel, constraining themselves to one particular type of roofs. This is not true in general, as seen in \myFigRef{fig:Geographic_locations}.
\cite{Ghaffarian2014} proposed the variation of the FastICA algorithm and k-means to detect buildings in monocular high-resolution Google Earth Images using the LUV space. Although their source of imagery is similar to us, they are relying on low-level and hand-crafted features and their objective is to achieve only built-up area segmentation and not the estimate of the number of buildings. 
  
Another direction is to group the low-level features intelligently for detecting buildings. 
\cite{krishnamachari1996delineating} used Markov Random Fields to detect buildings by combining the straight line segments detected from the edge map of an aerial image. In designing their objective function, they used the insight that only line-segments near each other needs to be combined and such combination should encourage rectangular shapes. 
\cite{ok2013automated} performed multiple graph-cuts using the shadow-cues to detect buildings in Very High-Resolution multi-spectral images.
Most of the previous works in building detection relied on the shadow detection \citep{ok2013automated, muller2005robust, huertas1988detecting, Ngoshadow-7579559, irvin1989methods} or shadow cues \citep{shadow-based}.

However, shadows depend upon the position of illumination source at the timing of the image capture. 
Several research works use more than just one optical Sensor \citep{Ngoshadow-7579559, ok2013automated} and rely on multi-spectral and$/$or the high-resolution satellite imagery.
Such systems mostly end up in segmenting the built-up area and fail in the cases where buildings are connected very close to each other.

Closest to our work is by \cite{xia2018extraction}, where authors segment out building instances, but they are using high-resolution imagery and difficult to get mobile LiDAR dataset. 
We solve the problem of counting the number of buildings, from satellite or aerial imagery in RGB space. 
This is a difficult problem, especially for densely populated areas in general \citep{Facebookzhang2017building}, or more specifically where the buildings are connected. The architectural and cultural designs impact how buildings appear from above, making it difficult to separately identify the boundaries of each building. 

 Counting objects from the images or videos \citep{idrees2013multi} is an interesting and important problem. 
 However, most of the recent works have been targeted towards the crowd counting, perhaps, because the dataset preparation for such is easier or the problem could be relegated to the counting of heads. Whereas the building does not have any such distinctive sub-part like a head.
Counting objects could be a complex or easy problem depending upon the sample. If objects are separable, a simple method is to detect objects and count them, for instance, see  \citep{doneObjectCountICCV2017, Learntorank, switching-8099912}. Furthermore, the recent success of deep learning based object detectors \citep{fastrcnn-7485869, yolo-7780460, Hu_2017_CVPR}, allows objects detection based counting methods to be more accurate.  
Moreover, many of these work exploits the structure of the object, for example, in crowd counting \citep{idrees2013multi}. \cite{Hu_2017_CVPR}, and \cite{CountingTinyFaces} uses the head shapes which are consistent across humans to detect heads and use that for the counting.
They also use perspective information i.e., the density of humans per pixel in patches far away from the camera will be more than the density of humans in patches near the camera. Many research works use the fact that the humans will be standing up and will be straight (in a way aligned to the axis). 
Data-set collected by the \cite{marsde2018people} also has the different densities of perspective properties. The same is true for the car counting problem. In sharp contrast, perspective information is not useful for satellite image building counting. Especially in the case of irregular construction, where houses of all sizes are build next to each other. 

DecideNet \citep{decideNet} comes closest to our work, in terms of trying to find a middle path between the count by detection and regression. However, that's where the similarity ends. Their algorithm relies on the object detection pipeline based on detecting the heads. Furthermore, their regression pipeline requires that the dots be placed on the heads of the humans. Both of these conditions are not applicable to our problem. There is no such visible "head" in terms of buildings and our pipeline does not require individual dots be given for each house.
Our method relies on exploiting built-area segmentation (not to be confused with instance level segmentation) for the attention.
 It only requires the total count concerning the image and not the sub-image count information as in previous counting methods. 

 Datasets play a vital role in the research, development, and evaluation of new technologies. \cite{lu2017exploring-8240966} proposed remote sensing captioning dataset, where each image is accompanied with five sentences. \cite{AID_Dataset-7907303} proposed a new large-scale scene classification dataset which includes 30 scene classes such as beach airport, desert, and farmlands. Other scenes classification datasets include UC-Merced \citep{UC-MercedDataset},  {NWPU-RESISC45 \citep{cheng2017remote}}, WHU-RS \citep{WHU-RS} and RSSCN7 \citep{RSSCN7-7272047} datasets. \cite{PatternNet2018} presented a 38 classes dataset for remote sensing image retrieval applications. 
 Finally, \cite{DrShah} put forward a new dataset for cross-view image geo-localization. In contrast, we collect a new challenging dataset that captures built areas of various densities from the satellite view.
 
 \begin{figure*}[t]
\begin{center}
        \includegraphics[width=2.0\columnwidth]{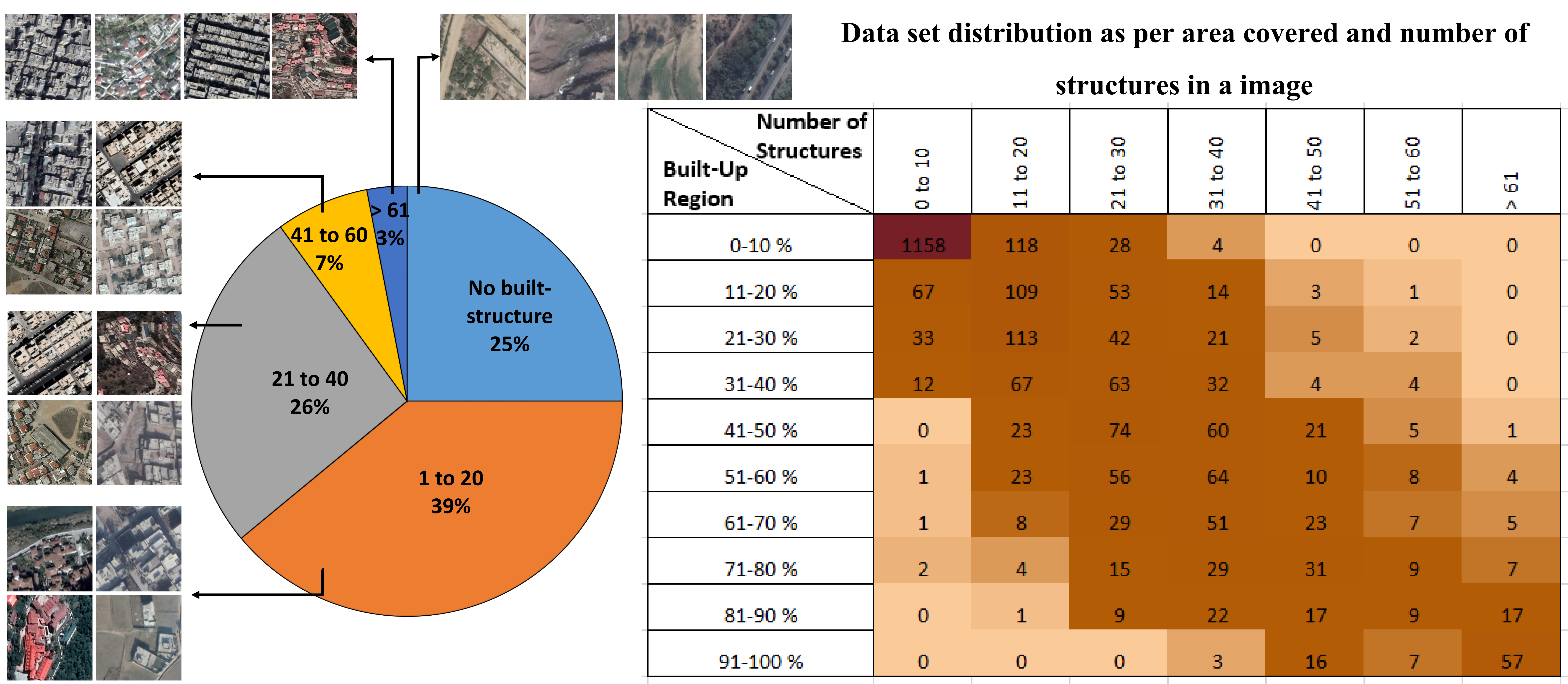}
    \caption{Distribution of data set according to the area covered by detected regions. (Left) A pie chart shows how the data is divided into count windows. Majority of the images in the data set corresponds to count window of 1 to 20. (Right) The built-up region is plotted against the number of structures. The covered area is divided into bins and number of images in each bin as per its count is computed hence represented by a heat map shown at right.}
    \label{Distribution_counts}
\end{center}
\end{figure*}

\section{Data Collection}
\label{sec:DataCollection}
The proliferation of deep learning libraries has enabled many to train for classification and regression tasks on the basis of the hyper or multi-spectral images without explicitly hand-designing, weighing, and fusion functions of different channels.
However, utilization of the deep-learning based methods in remote sensing has been challenged by the absence of the large-scale datasets \citep{demir2018deepglobe}.
 To the best of our knowledge, there is no publicly available dataset for counting buildings using satellite imagery, covering different geographical areas and a variety of built-up densities.
 Therefore, we have collected a new geographically diverse dataset by extracting, sorting and marking the satellite images.
 Although we mainly focused on counting the number of buildings, our dataset can be used for other remote sensing applications as well, such as more accurate surrogate for the population density estimation or neighborhood type estimation. 
 With the development of the latest high-quality hyper-spectral optical sensors, good quality high-resolution satellite images are publicly available for several developed countries. 
 However, still, the majority of publicly available satellite images are of low-quality as shown in Fig. \ref{fig:Geographic_locations} in comparison to the ground image datasets available today. 
 In this paper, we focus on RGB-images of a resolution (\textit{m per pixel}) which might be considered as VHR with respect to the satellite imagery, however, is of low quality when we consider sharpness of edges or noise, especially for the task on hand that is counting the number of buildings. 
 
 Note that the same building looks quite different depending on the position of the satellite and the time of the day the image was taken. 
 The height of the building, shadows, degree of separation and types of boundaries between the buildings makes these images challenging.
 To make our dataset realistic, we have collected satellite images at different times and from geographically different locations depicting built-areas of different diversity.
 Below, we provide details about our dataset collection and annotation process.

\subsection{Sorting and collecting data}
The satellite images are collected from regions including geographical and architectural differences that cover natural, urban and desert landscapes,  Fig.~ \ref{fig:Geographic_locations}. 
Collecting images from different locations induces scene-type variability and makes dataset challenging to evaluate. 
We selected different regions from these geographically different locations. From those regions, we have downloaded highly dense, moderately-dense and low-dense areas using Google Earth API. All these images are captured at zoom level 19 that covers 0.3\textit{m per pixel}. A building in a densely populated urban residential area covers approximately $25$-$30$ $m^2$, while in hilly regions and other rural areas, the range of area covered decreases to $15$-$20$ $m^2$.The tile size of $100\textit{m}\times100\textit{m}$ is selected on Google maps to capture all types of small, medium and large built-structures. The downloaded image size is $336 \times 336$. Table \ref{tab:Table_Kilometer} shows the details of the number of images downloaded from different landscapes with their areas in kilometer square. 

\begin{table}[h]
\begin{center}

\begin{tabular}{|l|c|c|}
\hline
Landscape   & Number of Images & Area Covered $km^2$  \\ \hline
Urban Areas & 2211             & 22.55         \\
Hilly Areas & 251              & 2.56          \\
Desert      & 220              & 2.24        \\ \hline
Total       & 2682             & 27.35        \\ \hline
\end{tabular}
    \caption{\small{Details of collected dataset.} \label{tab:Table_Kilometer}}\vspace{-0.4cm}
    \end{center}
\end{table}

Manually downloading geo-located images from Google maps is a daunting task, therefore a Matlab® based tool is designed that calls the Google Earth API to automatically download geo-located images. 
Specifically, at given size and scale, the image array and its corresponding latitude and longitude vectors are saved. Note that this pixel-level geo-location is very useful for visualizations and post-processing purposes.  
 
 Table in Fig. \ref{Distribution_counts} shows how challenging the collected dataset is. 
 The built-up region is computed using the satellite segmentation network explained in later sections. The percentage of area covered by the detected built-structures is compared with the labeled count of buildings. The comparison between both is made on the varying size of structures. As the satellite images are collected from various locations, the dataset covers a variety of different architectural designs of buildings with varying sizes that are difficult to learn from. As shown in the table in Fig. \ref{Distribution_counts}, there exist images containing few buildings but cover nearly $50-80$ \% of the area.
 

\begin{figure*}[h!]
       \includegraphics[clip, trim=1.0cm 0.1cm 1.0cm 1.0cm, width=16.5cm, height=6.5cm]{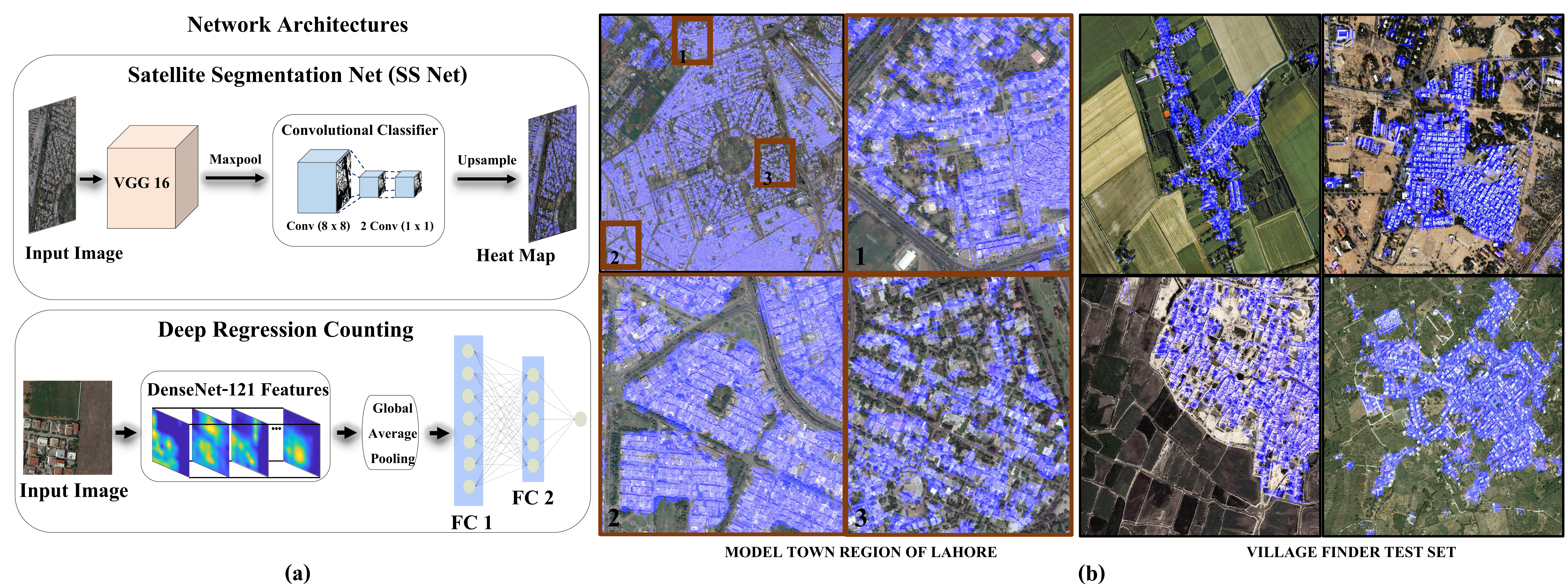}
    \caption{\small{\textbf{(a)} Network architectures, \textbf{Top:} pixel-wise classifier (SS-Net) to detect built-structures,  \textbf{Bottom:} Counting by regression model based on DenseNet features. Pooled features are fed into a three-layer neural network. \textbf{(b)} The qualitative results of SS-Net are presented on two test sets. \textbf{Left:} Building Detection probabilities overlaid on the region of the city of Lahore (Pakistan). The shades of blue represent the probable existence of built-structures, dark blue means high probability and light blue means low probability. Selected cells from the tile of Lahore, numbered from 1 to 3, are zoomed to show the accuracy of our algorithm. \textbf{Right:} Results on four of the samples from the village finders test set. }}
    \label{segmentation_results}
\end{figure*}

\subsection{Data tagging} \label{sec:Data tagging}
\vspace{-0.2cm}
Thorough annotation of the collected dataset was performed.   
 Specifically, we designed a Matlab based GUI to tag ground truth building count. To ensure good quality annotation, each image was annotated by at least two annotators. 
In Fig.~\ref{Distribution_counts}, we provide the detailed statistics of our dataset. The Pie chart on left shows the percentage of data that belong to a specific count window. As it can be seen that our dataset contains images with varieties of house count; from no built structure to a large count of built-structures. The Table on the right of 
Fig. \ref{Distribution_counts} shows the number of images in our data with a specific percentage of area covered by structures relative to the number of buildings in them. Note that we obtain built-up ratio using our Satellite Segmentation Net (Sec.~\ref{methodology_ssnet}).

\begin{figure*}[t]
\begin{center}
        \includegraphics[width=1.8\columnwidth]{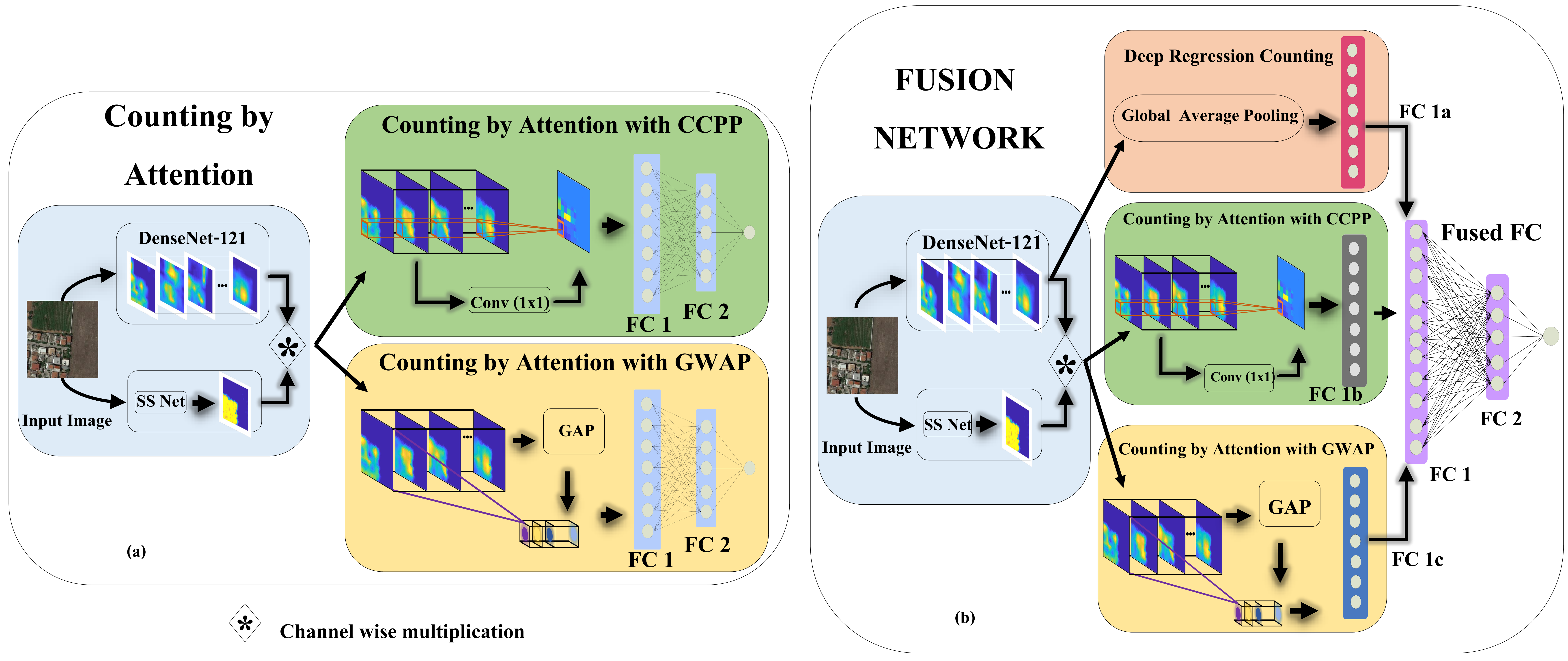}
    \caption{\small{The network architectures for the two proposed methods of counting by attention are shown. \textbf{(a)} The backbone (light-blue) is shared by both streams green and yellow. Channel-wise multiplication is performed between the probability map of SS-Net and the feature volume of DenseNet-121. This weighted feature map is input to both streams which are trained separately. The green block represents CCPP and the yellow one represents the GWAP. \textbf{(b)} Network architecture for FusionNet. Information captured by each stream is fused together to perform regression. The input of top stream (DRC) is the feature volume from DenseNet, while the CCPP and GWAP stream takes the weighted feature volume as input. FC layers from all three streams are concatenated together to form a fused FC and thus the error is back-propagated collectively.}}
    \label{attention_model}
\end{center}
\end{figure*}



\section{Methodology}\label{sec:Methodology}
The primary goal of our paper is to achieve a precise count of the number of buildings in each satellite image, which is a challenging problem, as usually, satellite imagery is of low-resolution and quality as compared to the generally available ground imagery.
Most importantly, there is no visible space between the neighboring buildings making it difficult to delineate accurately each building. 
Therefore, we propose to map deep visual features to real numbers representing the count of built-structures in the image. 
As an input to our regression model, initially, we took deep features from DenseNet \citep{DenseNet-8099726} and map them to house counting problem through a fully connected neural network. 
Although, we achieve decent counting results (see table \ref{mis-pred-ranges}), however, the DenseNet gives equal importance to all of the image regions, therefore, resulting in the loss of accuracy. 
Our key intuition is that for the house counting purpose, features belonging to built-up regions are more important than the features originating from the rest of the image regions such as form fields, streets etc. 
Therefore, we propose two deep regression approaches using attention based re-weighting (ABW), where we decrease the influence of deep features from non-built areas such as fields or streets regions; thus enabling the algorithm to predict count with more accuracy.
Our experimental results validate our intuitions. 
Below, we provide details for each of our proposed approach. 

\subsection{Deep Regression Counting (DRC)}\label{sec:DRC}
\vspace{-0.2cm}
We pose built-structure counting as a deep regression problem, that is, training deep learning based models with the regression as an output layer. 
Transfer learning \citep{Bengio:2011:DLR:3045796.3045800} is performed by extracting the deep features from global average pooling layer of DenseNet \citep{DenseNet-8099726}, pre-trained on ImageNet. 
Note that ImageNet \citep{krizhevsky2012imagenet} is a large dataset with 1K class labels.
Many recent works like \citep{Huang2017TransferLW} indicate that features learned by CNN models, e.g. VGG \citep{VGG} and AlexNet \citep{Krizhevsky}, trained on such large datasets can be used to perform transfer learning for tasks with limited training data. 
DenseNet was used because of its reported high accuracy and computational efficiency.

Features extracted are fed into the fully-connected (FC) neural network. 
We used a three-layered network having 512, 32, 1 units respectively
 (Fig. \ref{segmentation_results}).  
We used 60\% Dropout layer between FCs.
Relu is used as an activation layer after the first and second fully connected layer.
No activation function is applied at the output layer. 
We have not used the ImageNet mean values to normalize our remote sensing data. 
Though initializing weights with such datasets is helpful instead of random values but the mean of ImageNet is a pure representation of day to day ground images. In our experiments, normalizing satellite imagery using these values disrupts the input and this affects the accuracy of the model. 


\subsection{Deep Regression Counting by Attention}
\vspace{-0.2cm}
Deep Regression counting suffers from the problem of giving equal weight to all the features whether they belong to a built-up region or not. 
Attention-based architectures help neural network concentrate on the task at hand and not impacted by the noise. 
 To exploit local information for precise building count, we propose to use built-up region segmentation probabilities as the attention. 
 
\subsubsection{Satellite Segmentation Net (SS-Net):}\label{methodology_ssnet} 
 
 We train compact VGG-based \citep{VGG} fully convolutional neural network \citep{FCN-7298965} to perform pixel-wise built-up region classification. We call this network SS-Net. 
 The output convolutional layer of this network predicts if a 64$\times$64 input patch belongs to a built-structure or not. 
 The SS-Net is trained on low-resolution Village-Finders dataset \citep{VillageFinder}. The original size of the images in data is $512\times512$. We randomly crop the patches of size $64\times64$ and $128\times128$ from the image and use segmentation mask associated with them to generate labels. 
 The weights of the network are initialized with pre-trained VGG network (trained on ImageNet) weights and the data is normalized by computing its mean values instead of the ImageNet ones.
 During training, each patch in the training set is augmented four times by flipping, inverting and rotating the patch 45 degrees.
 Inspired by \citep{Bootstrap} and \citep{NegMining}, to cater to the problem of unbalanced data, a bootstrap technique to do hard negative mining method is applied. 
 Specifically, after every 15 epochs, new samples were evaluated and all with a false positive response were added as negative examples of the training set. Fig. \ref{segmentation_results} shows the network architecture of the SS-Net.
 
 \begin{table}[h]
    \centering
        \begin{tabular}{|c|c|}\hline
             \textbf{\small{Evaluating Metric}} & \textbf{\small{SS-Net results}}\\\hline
             Pixel-wise accuracy & 0.947\\
             F1 score & 0.8\\\hline
        \end{tabular}
    \caption{Segmentation results of SS-Net on Village Finders test set. The results demonstrate high accuracy of propose technique.}
    \label{test_results_segmentation}
\end{table}
{During inference, when we present $64\times64$ patch to the SS-Net, it returns the probability of this patch containing building(s) or part of the building.}
 Since SS-Net is fully convolutional, it is capable of processing images of any size greater than  $64\times64$ pixels. The following equation gives the output size of the feature map at any layer $n_{out}  =\frac {n_{in} + 2P - K } {S} + 1,$
where $n_{out}$ is output size of feature map, $n_{in}$ is size of input feature map, $P$ represents padding, $K$ shows filter size and $S$ represents stride.  In our experiments, we used $P=1$, $S=2$ for max pooling layers and $P=1$, $S=1$ for convolution layers, and the value of $K$ depends on convolution layers. For instance, for an input image of $224\times224$, after 3 max pooling layers of stride 2 and kernel size 2, and following convolution layer of filter size $8\times8$ and $1\times1$, we obtain probability maps of $21\times21\times2$, where 2 represents number of channels.
A probability map, $\mathcal{P}$, representing the input image is generated for each image and bi-linear interpolation is performed to re-size the map to that of input image size.
Qualitative results in Fig.\ref{segmentation_results} demonstrate that our SS-Net can segment the built and non-built areas with very high accuracy.
Table \ref{test_results_segmentation} demonstrates the accuracy and F1-score of SS-Net on village finders test set.

The building probability calculated on each pixel is used to improve the regression algorithm for counting buildings. 
In sections below, we discuss in detail our two proposed approaches that use output probability maps of SS-Net for improved building counts.


\begin{table*}[t]
\centering
\begin{tabular}{|l|c|c|c|c|}
\hline
Models                                                                           & \begin{tabular}[c]{@{}c@{}}\textbf{DRC}\end{tabular} & \begin{tabular}[c]{@{}c@{}}\textbf{GWAP}\end{tabular} & \begin{tabular}[c]{@{}c@{}}\textbf{CCPP}\end{tabular} & \textbf{FusionNet} \\ \hline
\begin{tabular}[c]{@{}l@{}}Total Absolute Error (Low-Count : 0 to 30)\end{tabular}           & 1158                                                                & 1121                                                             & 1136                                                             & 1001       \\ \hline
\begin{tabular}[c]{@{}l@{}}Total Absolute Error (Medium-Count:31 to 60)\end{tabular}       & 814                                                                 & 820                                                              & 796                                                              & 743        \\ \hline
\begin{tabular}[c]{@{}l@{}}Total Absolute Error (High-Count: $>$ 60)\end{tabular} & 229                                                                 & 180                                                              & 161                                                              & 176        \\ \hline
Total Absolute Error (TAE: Total)                                                           & 2201                                                                & 2121                                                             & 2093                                                             & 1920       \\ \hline

\begin{tabular}[c]{@{}l@{}}Mean Absolute Error (TAE$/$(Total Number of Images)) \end{tabular} & 4.14                                                                & 3.99                                                              & 3.94                                                              & 3.61        \\ \hline
\begin{tabular}[c]{@{}l@{}}$R^2$ (coefficient of determination) \end{tabular} & 0.86                                                                & 0.872                                                              & 0.875                                                             & 0.88        \\ \hline
\end{tabular}
\caption{\small{Total Absolute Error of structures in the set of Low-Count, set of Medium-Count and set of High-Count ranges, numbers in the bracket represent the building count in that satellite image patch. Where set Low-Count contains 3880, Medium-Count contains 3937 and High-Count contains 1128 structures in the test set. Mean Absolute Error (MAE) and $R^2$ score of each model is also listed.}} \label{mis-pred-ranges}
\end{table*}

\subsubsection{Global Weighted Average Pooling (GWAP):}\label{GWAP}
Similar to Sec.~\ref{sec:DRC} the pre-trained DenseNet is used to extract the features. 
However, in this algorithm attention map generated by the SS-Net is used to perform Attention Based Global Weighted Average Pooling (GWAP) over the features.
To achieve GWAP, we first multiply each feature map, extracted from the DenseNet, with probability map generated by SS-Net and then compute the average of each channel independently.
This results in dimensionality reduction while gathering of the spatial information.
Each value in the pooled vector corresponds to the density of constructed regions in a satellite image. 
These activation maps and SS-Net output probability maps for a typical image are shown in Fig. \ref{attention_model} under the 'Counting by Attention' pipeline.
Similar to GAP \citep{NetworkInNetwork}, GWAP also directly corresponds to the features learned. 
However, in GWAP features from different locations of the image are given different weights.
As shown in Fig. \ref{attention_model}, DenseNet produces features maps (output of last convolutional layer) which are agnostic to the built structure while SS-Net provides high probability score on the built area. Combining these two maps filters out the activation values of DenseNet from non-built areas.
This meaningful representation is then fed into the 3-layer fully connected neural network, with 512, 32 and 1 units referred as regression pipeline.
The yellow block along with blue block (Fig.~\ref{attention_model}) displays the network architecture of GWAP. 

\begin{figure*}[h]
\begin{center}
    \includegraphics[clip, trim=3.0cm 10.3cm 1.0cm 0.0cm, width=0.95\textwidth]{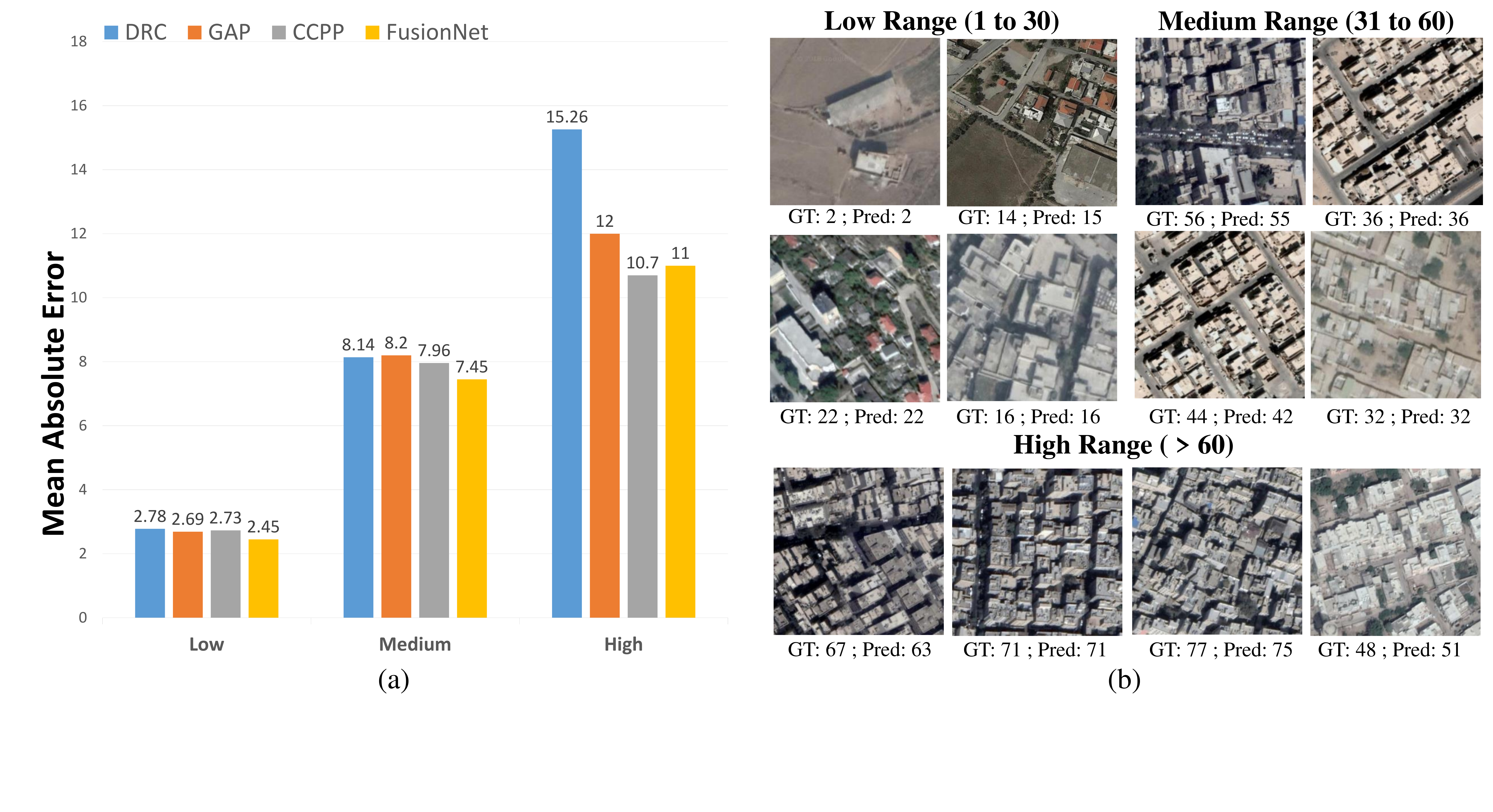}
    \caption{\vspace{-0.5cm} 
\textbf{(a)} Mean absolute error of all categories. \textbf{(b)} Ground truth and predicted count of the testing samples inferred using FusionNet. }
    \label{results1}
\end{center}
\end{figure*}
\subsubsection{Cross Channel Parametric Pooling (CCPP):}\label{CCPP}
GWAP allows the algorithm to consider only the built area, however, it suffers from a lack of accuracy.
One reason is the effect of averaging i.e. features representing buildings at different locations are summed up. 
However, recognizing them separately is required for accurate building count, especially for the densely built-up areas.
Instead of predicting one single value for the whole image, if one can predict the count at different locations of the image, then the final number should be a summation of these counts; thus reducing the effect of averaging.
However, we only have one count per image and not the count at each location. 
To counter this shortcoming, we design a network that can take care of across the channels correlation and spatial layout of the feature map.
Specifically, we employ convolution of kernel size $1\times1$, let's call it $C_1$ that outputs a single activation map. 
This activation map is presented to the fully connected regression pipeline, predicting the final count.
Note that the architecture of regression pipeline is same in all methods, however, minor changes regarding the activation function and optimizer were experimented and are discussed in the implementation details.

The output of the layer $C_1$ is a single channel, visualized in the green block (Fig.~\ref{attention_model}). 
This convolutional layer $C_1$ performs learnable interactions within the weighted feature volume at every location. This layer learns the combination and comparison of all sizes of built-structures that are captured by the weighted feature map. 
Its response is different at different parts of the images, corresponding to the density of the buildings at that location.

\subsubsection{Counting by Attention with FusionNet:}
All the models discussed above suffer from one or other shortcoming. 
GWAP is unable to give credence to the local information. 
CCPP, where handles the local information, is challenged when the images with low count are presented, much due to the lack of the larger perspective (Table \ref{mis-pred-ranges}). 
The attention based pipelines (Sec.~\ref{GWAP} \& \ref{CCPP}) do better than generic deep regression pipeline in our case, by detecting the areas with buildings.
However, as shown in Fig. \ref{segmentation_results}, our building segmentation system takes away the other useful information too, such as the location of streets or roads, or other markers highlighting the natural boundaries of the buildings. 
FusionNet has been designed to counter the shortcomings and enhance the benefits, by fusing the features extracted from each method. 
These fused features are processed by the fully connected regression network, outputting the final count.
After concatenating the output of FC layers, the number of units in the fused layer is 1536. Finally, the fused layer is fed into the 3-layer fully connected neural network, with 512, 32 and 1 units referred before as regression pipeline. The network architecture of FusionNet is displayed in Fig. \ref{attention_model}.
 
 All above approaches when fused together complement each other hence improving the learning of regression pipeline. 
 During training, the penalty is back-propagated collectively where all or any of the streams results in the prediction of an erroneous count.
     

\subsection{Implementations details}
\vspace{-0.2cm}
All regression-based models are trained on $336\times336$ image size in pixels which correspond to $100 \textit{m}\times100 \textit{m}$ area covered on the ground with a resolution of 0.3 \textit{m per pixel}. 
While using DenseNet features, we did not use any normalization technique as per our experiments, normalizing remote sensing data with ImageNet mean-values disrupts the images.
To prevent the model from over-fitting, Dropout layers \citep{dropout} are applied with a ratio of $0.6$ on fully connected layers. 
Apart from FusionNet, all models have the same regression pipeline comprising of three FCS of 512, 32 and 1 units. 
In FusionNet, last fully connected layer of all three blocks (DRC, CCCP, GWAP) contains 512 units. 
Concatenating them creates an input layer of 1536 dimension, which is input to fully connected layer of 512 units, followed by 32 units and 1 unit fully connected layers. While training the Deep Regression Counting, we use ReLu as an activation function. 
However, for all attention based models, leaky-ReLu with a ratio of $0.3$ is used. This counters the high activation values resulting from the DenseNet features and its product with the probability maps.
For training the built-up area segmentation network, we normalize by subtracting the mean of the whole train set from each image. SS-Net is trained with a batch size of 16 on patches of size 64, so only the first three blocks of the VGG-16 are used. The learning rate of \(1e^{-5}\) is used with optimizer SGD to train the SS-Net. 
The training data, for counting, was augmented by flipping, inserting images, and rotating them at angles 90 degrees and 270 degrees, increasing its size five times. All the experiments were performed using Keras with tensorflow as a backend.
{Chanel-wise cross-entropy loss function is used for training SS-Net. For training DRC, mean squared is used. Furthermore, all attention based training networks (GWAP, CCPP, and FusionNet) are trained on root mean squared error.}

\begin{figure}[t]
\begin{center}
        \includegraphics[width=1.0\columnwidth]{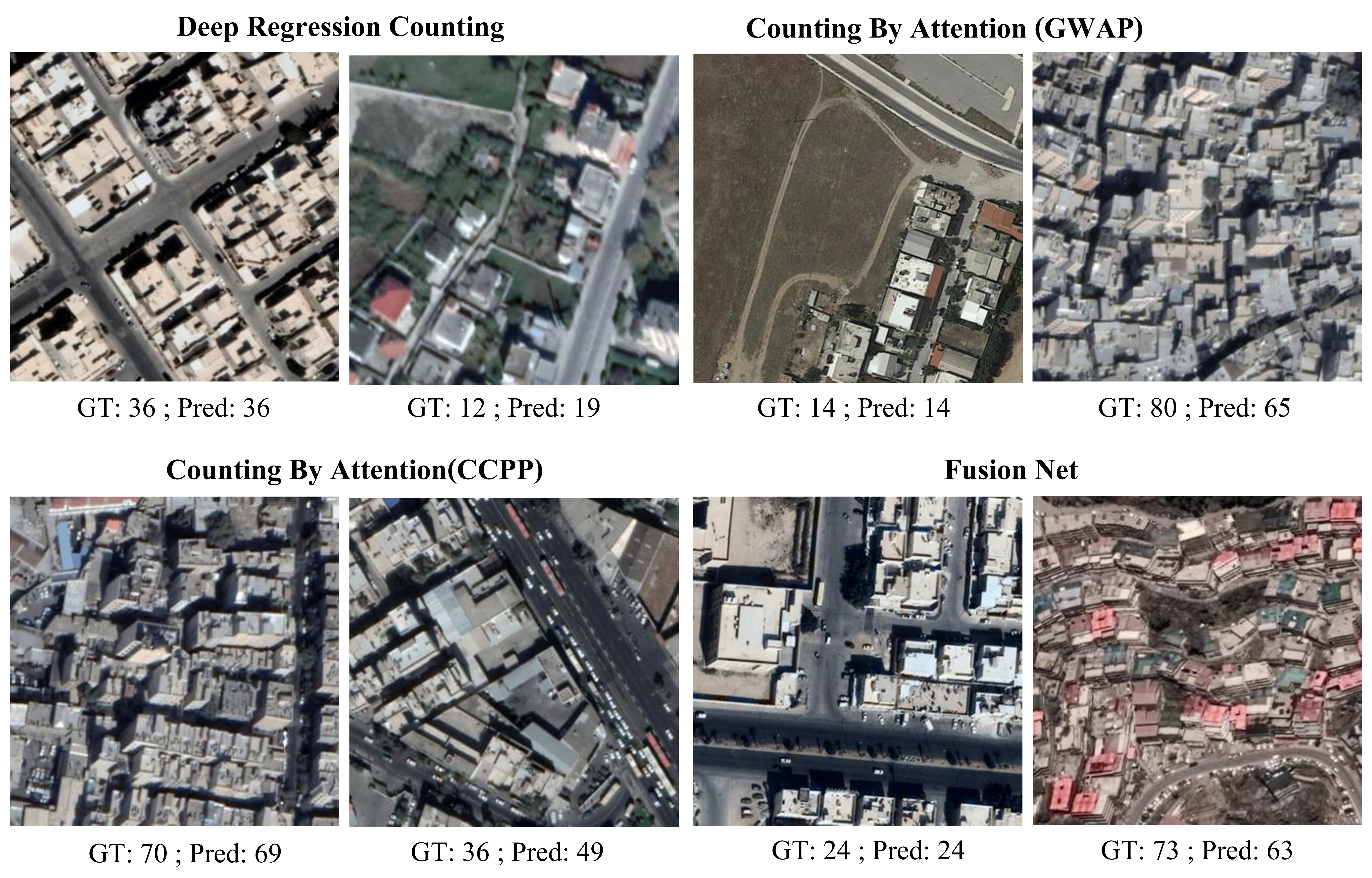}
    \caption{For each method two sample results are shown, one where our prediction is accurate and one where it's not (Sec. \ref{sect:analysisResult}).}
    \label{results2}
\end{center}
\vspace{-0.5cm} 
\end{figure}

\section{Results and Analysis}\label{sec:ResultsandAnalysis}
\vspace{-0.2cm}
A thorough comparative analysis of proposed approaches is performed by evaluating their results on the test set of 531 satellite images, extracted from our collected dataset.
In Table \ref{mis-pred-ranges}, we provide a quantitative comparison of all four proposed approaches, by calculating the mean absolute error (MAE), total absolute error (TAE) and R-squared measure.
The MAE decreases and R-squared values improve, as we move from deep regression counting to FusionNet.

In order to perform in-depth analysis of our results, the test set is divided into three ranges on the basis of ground truth count of the buildings; \textbf{(a)} Low-Count (0 to 30): less built, \textbf{(b)} Medium-Count (31 to 60) : reasonably populated and \textbf{(c)} High-Count (greater than 60) : densely built. 
Out of 531 images 416, 100 and 15 are in the Low-Count, Medium-Count and High-count range, respectively.
TAE for all four approaches on each set is computed separately. The 531 testing images cover a total of 8945 structures. 
MAE is computed by dividing the total absolute error with the total number of images. 
As compared to deep regression counting, both attention re-weighted counting has better results. Finally, the fusion of all three approaches further decreases the mean absolute error (see Table \ref{mis-pred-ranges}). {The proposed approach is quite efficient; DRC, GWP and FusionNet took 0.07 (0.02), 0.8 (0.026), and 0.9 (0.029) sec/image (sec/Km) respectively. }

\subsection{Comparison and Analysis of results }\label{sect:analysisResult}
\vspace{-0.2cm}

As indicated by the MAE results, Table \ref{mis-pred-ranges} and Fig. \ref{results1}, introducing attention mechanism considerably decreases the MAE (3.6\%) and increases the R-squared value. 
Fig. \ref{results2} shows the images corresponding to minimum and maximum MAE, for all of our proposed models. 
On fine-grain analysis, it is observed that the GWAP network is accurate for the Low-Count images whereas the CCPP network is predicting with lesser TAE in the Medium-Count and High-Count images (Table \ref{mis-pred-ranges}).
For the low-density images, where both CCPP and GWAP are much better than DRC, GWAP's TAE is much less than that of CCPP.  
With the involvement of attention the MAE between the ground truth and predicted count decreases generally but the CCPP seems to be distracted while over counting the structures in some of the images. For example, in   Fig. \ref{results2}, vehicles parked on the road are misleading the model.
However, for both medium and high-density images, the number of TAE of CCPP is much less than the GWAP, indicating that much more detailed local information is needed for counting where the density of buildings is more. 
To capitalize on the complementary nature of CCPP and GWAP, FusionNet is trained which combines the deep regression counting with both attention models. Fig. \ref{results1} shows the comparison in the MAE of these models. 
{
We retrained DRC on the mean (of ImageNet) subtracted data. MAE of this mean-subtracted DRC rose from 4.14 to 16.98.}
{ Deep regression was performed on the features extracted from SS-Net. This resulted in an increase in MAE to 5.5 since these features do-not capture inner-structure in the segment}

\subsection{Counting in large neighbourhood}
In order to show generalization capacity and effectiveness of our model, we test our approach on a portion of Cairo's densely populated region. 
The satellite image is of the size $1008\times3024$ pixels, covering $302.4\times907.2m^2$ area. 
The region covered in this testing tile is diverse, containing both small and large structures in densely and moderately populated areas.
Ground truth is created by manually counting the buildings, and came up to be 656.
Our approach, FusionNet predicted 675 buildings which are quite close to the ground-truth.  
In order to perform detailed analysis, we divide the image into 27 cells, where each cell is of size $336\times336$ pixels. 
FusionNet's prediction for each of the cell is compared with the ground-truth count of the buildings in that cell, the ground-truth is achieved by hand-marking each cell in the image.
Predicted count is overlayed on the map for visualization, Fig. \ref{large_image}, by assigning different colors according to the different predicted counts in each window. 
For a quantitative comparison, graph of predicted count and ground truth count is show in  Fig. \ref{large_image}, indicating that predicted values closely follow the ground-truth.
We argue that cell 18 is intensely populated and contains irregular construction which makes it difficult even for the human annotators to count.
High accuracy on a large image outside of training and test set, demonstrates the generalization capacity and robustness of our proposed approach. 

\section{Conclusion}\label{Conclusion}
 In this paper, we have attempted to solve a difficult problem of counting buildings from satellite imagery.
 The diversity in the shape of the urban structures, variations in city planning and sensor response, makes the problem challenging.  
 We have introduced a new challenging benchmark dataset capturing different geographical regions and areas with different building counts at various build densities (dataset will be made publicly available).
Instead of using deep learning as a black box, we have presented an attention based mechanism, based on insights of how Deep Convolutional Neural Networks work so that our model can capture variations in the urban-structures.
Our final solution, FusionNet, combines the information captured by different pipelines at different granularity, making it robust to the densely built buildings, as well as to sparsely built areas, from large structures (covering a large area) as well as to small structures. 
FusionNet is able to handle a variety of the roof types including a difficult case of flat roofs especially when the buildings are interconnected. 
   Future directions include improving the image quality through super-resolution before feature computations and investigation of other pooling techniques to improve building counting.\newline
   
\noindent{\textbf{Acknowledgment:} We greatly appreciate discussion and useful comments provided by Hamza Rawal, Maria Zubair, Komal Khan and Umar Saif.}


{
    \begin{spacing}{0.9}
        \bibliography{bibliography} 
    \end{spacing}
}


\textit{Revised March 2019}
\end{document}